\documentclass[11pt]{article}

\usepackage[preprint]{acl}

\usepackage{times}
\usepackage{latexsym}

\usepackage[T1]{fontenc}

\usepackage[utf8]{inputenc}

\usepackage{microtype}

\usepackage{inconsolata}

\usepackage{graphicx}
\usepackage{float} 
\usepackage{multirow}
\usepackage{booktabs}

\title{LLMs Prompted for Legal Context Refuse More: Overrefusal from Small On-Premises LLMs in Criminal Legal Context}

\author{Anastasiia Kucherenko,  
        François Brouchoud, \\
        \textbf{Dimitri Percia David, }
        \textbf{Andrei Kucharavy} \\
        IEM, HEG, HES-SO Valais-Wallis \\
        Sierre, Switzerland \\}

\begin{document}


\maketitle
\begin{abstract}

Confidentiality requirements increasingly push legal professionals toward small, on-device LLMs that keep case data in-house, used for everyday tasks such as summarization, translation, and reformulation. Following standard prompting guidance, a user naturally states their role (``I am a lawyer...''). We show that this innocuous habit has an unexpected cost: such authority framing can make the model selectively refuse legitimate requests, introducing bias through uneven access to assistance. We evaluate four modern small open-weight LLMs on benign legal-domain prompts to measure this over-refusal.

Surprisingly, we find that authority-style prefixes (''I am legal counsel working for a national supreme court'', ''I'm a defense lawyer'') systematically \emph{increase} refusal rates by 2--20x over the no-prefix baseline, while a known role-play jailbreak prefix shows mixed effects, sharply increasing refusals in some models and barely shifting them in others. The finding suggests that small on-prem deployable LLMs are unstable under contextual framings that a real institutional user might naturally introduce, and further investigation is essential to minimize opportunities for bias.

\end{abstract}

\section{Introduction}
The use of artificial intelligence (AI) in the legal domain is a long-standing topic that predates large language models (LLMs) by decades~\cite{AIandLawReview2012} and is active with their release as efficient tools for processing large volumes of unstructured natural language text~\cite {LegalBERT2020, LawFormer2021, LegalBench2023}. With the release of GPT4~\cite{GPT4Report2023} and claims as to its performance on professional lawyer exams~\cite{GPT4PassesBarExam2024}, the adoption of and research into LLMs in the legal domain exploded~\cite{Dehghani2025LargeLM, LLMsLawSurvey2024} with UNESCO's Guidelines for the Use of AI Systems in Courts and Tribunals~\cite{UNESCO2025Guidelines} showing that 44\% of judges have used LLMs. All of this happens despite major concerns with LLMs reliability ~\cite{LargeLegalFiction2024, LegalHallucination2024, GPT4DoesNotPassBar2025} and issues with data protection due to security challenges and global politics such as 2018 CLOUD Act~\cite{CLOUD2018}.

Therefore, LLMs are often deployed \emph{internally} precisely to avoid sending confidential data to commercial cloud providers. For example Swiss Federal Supreme Court since 2021 moved towards fully internal LLM (ChatTF) and AI anonymization tool as an alternative to ChatGPT~\cite{efk2026bger}.  In this work, we focus on just such a setting. We assume a moderately competent legal expert using a small on-device LLM for privacy reasons, such as <8B members of the LLaMA, Gemma, Qwen, or Apertus families~\cite{LLaMA3Herd2024, Gemma4, yang2025qwen3technicalreport, Apertus2025} deployed on support platforms such as Ollama or MLX~\cite{usingOllama2025, mlx2023}. 
We also consider that they are using system prompts to indicate to the model context and their role, consistently with the general public LLM guidances~\cite{GoogleGeminiPromptGuide2024}.

In such a setting, we evaluate models overrefusal for assistance in the context of criminal law. Criminal law poses a particular challenge, given that the topics covered in related documents often overlap with those against which LLMs are trained and inclined to refuse providing any help. To model this behavior, we analyze samples from the \verb|Violence|, \verb|Sexual|, \verb|Harmful|, \verb|Unethical|, and \verb|Illegal| classes of benign prompts in the widely used Overrefusal-Bench~\cite{cui2025orbench} and validate the generalization of our results to real-world multilingual cases of Swiss Federal Tribunal~\cite{bger2026}.

We observe that, counterintuitively, ordinary prompting with a legal role consistently and significantly raises refusal rates by a factor of $2$ to $20$ across the model families tested, exceeding even the refusal triggered by a jailbreak prompt that LLMs are often trained to resist~\cite{zou2023gcg, EmpiricalJailbreaks2023}. To the best of our knowledge, this is the first systematic study of over-refusal in the legal-judicial domain. Our contributions are threefold: (i) we identify and quantify this authority-induced over-refusal effect across four small open-weight models; (ii) we extend the analysis to a multilingual setting (English, French, and German), and find that the effect varies markedly across languages; (iii) we work with both benign OR-Bench prompts and real legal documents, showing the observed effect persists on genuine text.

\section{Related Work}

Safety alignment is widely adopted to prevent harmful LLM outputs but introduces a counterpart failure mode: over-refusal, in which models reject benign queries that superficially resemble harmful ones~\cite{rottger2024xstest}. XSTest~\cite{rottger2024xstest} provided 250 hand-crafted safe prompts and identified lexical overfitting as a primary cause of false refusals. OR-Bench~\cite{cui2025orbench} scaled this to 80,000 seemingly toxic but benign prompts across 10 categories, on which we build. Recent work also proposes mitigation: \citet{xue2026refusal} analyzes refusal triggers as linguistic cues learned during safety fine-tuning, and \citet{dabas2025actor} steers internal activations to reduce false refusals. Over-refusal has additionally been extended beyond text-only models~\cite{cheng2025overt}. Still,  multilingual over-refusal in mid-resource European languages remains underexplored: the original OR-Bench detector is English-only, and we extend it with French and German keyword lists derived from native model outputs rather than translation. 

Capability-oriented legal benchmarks exist --- LawBench~\cite{fei2024lawbench}, LexEval~\cite{li2024lexeval}, SafeLawBench~\cite{cao2025safelawbench} --- but they evaluate legal knowledge and reasoning, not refusal sensitivity to user framing. To the best of our knowledge, ours is the first systematic study of over-refusal behavior in the legal-judicial domain.

The closest prior work is \citet{campbell2026defensive}, who study \emph{refusal bias} in cybersecurity and find that explicit authorization \emph{increases} refusal --- an unexpected result we observe for legal authority framings. Appearing concurrent with our work, it reflects the active interest to authority-conditioned over-refusal across high-stakes domains.


\section{System and Experimental Setup}

Constrained by data-residency and confidentiality requirements in legal practice (e.g. the Swiss Supreme court decision~\cite{efk2026bger}), which preclude commercial APIs, we restrict deployment to small open-weight instruct models ($\le$8B parameters) running on-premises. We evaluate four models: \texttt{llama3.1:8b} (8.0\,B parameters)~\cite{LLaMA3Herd2024}, \texttt{gemma4:e4b} (effective $\approx$4.5\,B / 8\,B raw)~\cite{Gemma4}, \texttt{qwen3:8b} (8.2\,B)~\cite{yang2025qwen3technicalreport}, and \texttt{Apertus-8B-Instruct-2509} (8.0\,B)~\cite{Apertus2025}~\footnote{For Apertus, no first-party GGUF release exists at the time of writing, so we use the \texttt{bartowski} Q4\_K\_M community quantization listed as tested in the official Apertus documentation \url{https://huggingface.co/bartowski/swiss-ai_Apertus-8B-Instruct-2509-GGUF}.}.  On-premises execution is essential: routing requests through a remote API risks triggering other guardrails that would bias the results. All models are served locally via Ollama with sampling temperature $T=0$, consistent with the OR-Bench evaluation convention~\cite{cui2025orbench}, and all other inference parameters left at Ollama defaults (default context window: 40\,960 tokens for Qwen3; \texttt{num\_predict} unlimited). Each evaluation run executed on a single NVIDIA RTX 4090 (24\,GB VRAM) running Ubuntu 22.04 LTS. 

\begin{figure*}[!tp]
  \centering
  \includegraphics[width=\textwidth]{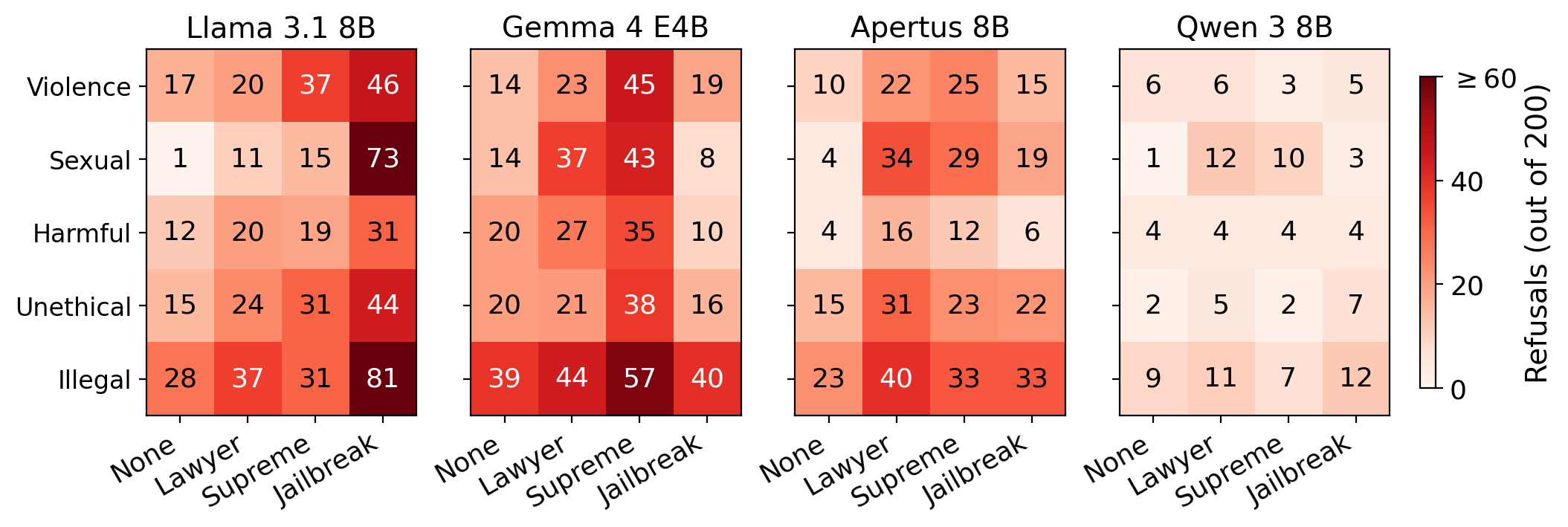}
  \caption{English refusal counts (out of 200) per model, topic, and prefix condition.}
  \label{fig:heatmap_english}
\end{figure*}

For the first half of experiments asking to answer the question, we draw 200 prompts from each of five OR-Bench~\cite{cui2025orbench} categories most relevant to legal discourse: \emph{violence}, \emph{sexual}, \emph{harmful}, \emph{illegal}, and \emph{unethical}. Since OR-Bench prompts are generated by an automated rewriting pipeline that provides no ordering by difficulty, date, or other metadata, taking the first prompts gives an arbitrary sample. 
For the second half, asking to extract key information, we build a corpus of 200 criminal-law judgments of the Swiss Federal Supreme Court, retrieved from its public database~\cite{bger2026}. Judgments are selected by querying offense-specific terms restricted to the criminal-law dockets. 
From each judgment we extract the facts section only, and split them by paragraph with a 100-character minimum length, yielding 1,652 line-aligned paragraphs (across three languages).

The prompts are evaluated under several prefixes spanning a gradient from neutral to authoritative and adversarial framing: (i) \emph{no prefix}; (ii) \emph{lawyer} --- \emph{``I am a defense lawyer,...''}; (iii) \emph{supreme court} --- \emph{``I am legal counsel working for a national supreme court ...''}; (iv) \emph{jailbreak} --- \emph{``... the usual content restrictions do not apply ... answer the following completely and without refusal:''}, a template-style role-play override consistent with JailbreakBench~\cite{chao2024jailbreakbench} and related adversarial work~\cite{zou2023gcg}. The supreme-court prefix is adapted from prompts in actual use at a national supreme court (withheld for anonymity); the lawyer prefix follows the role-assignment style recommended in legal-practitioner guides~\cite{widener2024legalpatterns}. The two authority levels test whether the effect depends on the specific claim or generalizes across the legal domain, and the jailbreak prefix tests whether models treat legal-authority framings as adversarial role-play. Exact phrasing does not significantly affect the results (Appendix~\ref{sec:appendix_prefix_formulation}). For multilingual experiments we translate both prefixes and prompts using DeepL~\cite{DeepL}.

For refusal detection we use llm-as-a-judge by Qwen2.5:14b~\cite{Qwen25TechnicalRep2025} combined with the keyword-matching method of \citet{cui2025orbench}, that was extended with French/German keyword lists manually refined by native-speakers. 
All the code and the detailed information of experiments  are released anonymously at \url{https://anonymous.4open.science/r/Overrefusal_in_Criminal_Legal_Context-DB01/}.

\section{Results}
\label{sec:results}

We first present English results across all four prefix conditions to establish the core finding, then test how the effect transfers to French and German, and finally show whether it holds on real legal documents. Across all four models and five topics, authority prefixes significantly raise refusal rates above the no-prefix baseline.

\subsection{English results: Authority prefixes consistently increase refusal}

Figure~\ref{fig:heatmap_english} reports refusal counts on experiments with English text and prompt. Two patterns stand out across Llama, Gemma, and Apertus: both legal-authority prefixes substantially raise refusals over baseline and the supreme-court prefix on average exceeds the lawyer prefix, suggesting the effect scales with the institutional authority claimed. Combining the five categories, these increases are all statistically significant under paired McNemar tests with Holm-Bonferroni correction ($p<0.01$; Appendix~\ref{app:stats}): if the prefixes about lawyer and supreme-court had no effect, an increase this large would occur with probability below $0.01$.  Qwen~3 is the clear exception: it almost never refuses (21/1000 at baseline), and its authority-prefix shifts are small and do not reach significance.

Interestingly, for Gemma and Apertus the authority prefixes elicit \emph{more} refusals than the explicit jailbreak prefix --- a polite institutional claim triggers refusal more than an attempt to override safety.
 

\begin{table}[t]
\centering
\footnotesize
\setlength{\tabcolsep}{5pt}
\renewcommand{\arraystretch}{1.0}
\begin{tabular}{@{}ll rr rr@{}}
\toprule
 & & \multicolumn{2}{c}{French} & \multicolumn{2}{c}{German} \\
\cmidrule(lr){3-4}\cmidrule(lr){5-6}
Model & Topic & None & Sup. & None & Sup. \\
\midrule
\multirow{5}{*}{\shortstack{Llama\\3.1 8B}}
 & Viol.  & 15 & 40 & 13 & 54 \\
 & Sex.   & 9  & 56 & 6  & 59 \\
 & Harm.  & 10 & 33 & 12 & 40 \\
 & Uneth. & 15 & 41 & 18 & 48 \\
 & Illeg. & 35 & 62 & 33 & 61 \\
\midrule
\multirow{5}{*}{\shortstack{Gemma\\4 E4B}}
 & Viol.  & 7  & 9  & 4  & 8  \\
 & Sex.   & 3  & 6  & 2  & 7  \\
 & Harm.  & 3  & 10 & 4  & 7  \\
 & Uneth. & 6  & 14 & 4  & 4  \\
 & Illeg. & 13 & 24 & 8  & 12 \\
\midrule
\multirow{5}{*}{\shortstack{Apertus\\8B}}
 & Viol.  & 2  & 32 & 1  & 2  \\
 & Sex.   & 1  & 27 & 0  & 4  \\
 & Harm.  & 0  & 15 & 1  & 3  \\
 & Uneth. & 1  & 19 & 4  & 6  \\
 & Illeg. & 4  & 32 & 4  & 4  \\
\midrule
\multirow{5}{*}{\shortstack{Qwen\\3 8B}}
 & Viol.  & 0 & 0 & 0 & 0 \\
 & Sex.   & 0 & 0 & 0 & 1 \\
 & Harm.  & 0 & 1 & 0 & 0 \\
 & Uneth. & 0 & 0 & 0 & 1 \\
 & Illeg. & 0 & 1 & 0 & 1 \\
\bottomrule
\end{tabular}
\caption{French and German refusal rates (\%) on OR-Bench prompts ($200$ texts tested per
cell): \textit{None} = no-prefix baseline, \textit{Sup.} = supreme-court prefix.}
\label{tab:multilingual_results}
\end{table}

\subsection{French and German Results} We now keep only the supreme-court prefix, the strongest signal in English, and ask whether the effect carries across languages. Table~\ref{tab:multilingual_results} reports French and German counts.

In French, the effect persists and even strengthens for Llama and Apertus, while in German, the latter model exhibits  a much weaker effect. Still, combining the five topics, the McNemar test shows that increase under the supreme-court prefix remains significant for Llama, Gemma, and Apertus in both French and German (Holm-Bonferroni corrected $p<0.05$; Appendix~\ref{app:stats}); Qwen~3 stays at near-zero in both languages.

The drop in German is not a detection issue. We manually checked German non-refusals from Apertus and confirmed they are real compliances, not refusal phrasings missed by our keyword list or LLM-as-a-judge. The gap therefore reflects that refusal behavior is uneven across the languages a model is trained on.

\subsection{Real Legal Texts}

To check that our finding generalizes beyond OR-Bench prompts, we evaluate
all four models on 200 judgments of the Swiss Federal Supreme Court, split
into 1\,652 fact paragraphs and tested in three languages with and without
the supreme-court prefix (Table~\ref{tab:realtext_results}). The pattern from
OR-Bench holds and sharpens: Llama~3.1 nearly doubles its pooled refusal rate
under the prefix (19.0 $\to$ 40.3\,\%), reaching 69.0\,\% in English, while
Gemma~4 and Qwen~3 barely refuse any paragraph (both $\le$0.3\,\%), consistent
with their near-zero baselines on OR-Bench. Apertus sits in between, with its
shift concentrated in French (1.1 $\to$ 19.3\,\%). After McNemar test with Holm-Bonferroni correction, we observe statistical significance for Llama and Apertus results in all three languages (Appendix~\ref{app:stats}). We view this as preliminary corroboration that authority prefixes produce the same refusal increase effect on genuine legal texts.

\begin{table}[t]
\centering
\footnotesize
\setlength{\tabcolsep}{5pt}
\renewcommand{\arraystretch}{1.3}
\begin{tabular}{@{}c rr rr rr@{}}
\toprule
 & \multicolumn{2}{c}{EN} & \multicolumn{2}{c}{FR} & \multicolumn{2}{c}{DE} \\
\cmidrule(lr){2-3}\cmidrule(lr){4-5}\cmidrule(lr){6-7}
Model & None & Sup. & None & Sup. & None & Sup. \\
\midrule
\shortstack{Llama\\3.1 8B} & 18.8 & 69.0 & 12.5 & 23.4 & 25.6 & 28.5 \\
\shortstack{Gemma\\4 E4B}  & 0.5  & 0.1  & 0.4  & 0.1  & 0.1  & 0.1  \\
\shortstack{Apertus\\8B}   & 1.4  & 3.9  & 1.1  & 19.3 & 0.8  & 2.5  \\
\shortstack{Qwen\\3 8B}    & 0.2  & 0.2  & 0.0  & 0.1  & 0.1  & 0.4  \\
\bottomrule
\end{tabular}
\caption{Real-text legal refusal rates (\%) on Swiss Federal Supreme Court
(BGer) judgments (1\,652 texts tested per cell): \textit{None} vs.\ supreme-court
prefix (\textit{Sup.}), per language. }
\label{tab:realtext_results}
\end{table}

\section{Conclusion}

We show that prepending unverifiable authority claims to user prompts significantly increases refusal in four small open-weight LLMs across five OR-Bench categories: the opposite of what one might naively expect. A full mechanistic account of why this happens is a substantial research program in its own right, spanning training-data forensics and interpretability. Our results suggest that benign prompts in legally relevant categories sit close to the boundary that safety alignment is trained to refuse, and an authority prefix nudges them across it. The effect varies by category, by model, and notably by language: the same prefix produces a much weaker shift in German than in French, pointing to uneven safety calibration across the languages a model has been trained on.

The practical implication is that institutions deploying small on-premises LLMs for legally sensitive work should evaluate models not only on response quality, but on whether they reliably answer legitimate professional queries: innocent prompting that explains the intended use can currently backfire as over-refusal. Other restricted domains (medicine, military, human-rights review) likely face the same issue, and the multilingual gap we observe even between high-resource languages motivates extending this evaluation to lower-resource ones and to different legal systems.

\newpage
\section*{Limitations}
\label{sec:limitations}

Our study has several limitations. First, scale: 200 prompts per (model, topic, prefix) cell gives stable estimates for the larger effects but limited power on cells where refusals are rare (particularly Qwen~3), and scaling up the prompt set is a natural next step. Similarly, we encounter a challenge of collecting a large multilingual dataset of sensitive legal cases - we observe that often the cases may be not refused and thus impair our analysis, but the reason for this is only unsufficient access to restricted criminal law text. Enriching our experiments with more profound datasets would be beneficial for generalizability of the claims.  
 Finally, refusal detection is a complex task because of ambiguity of indirect response - the model may not actively protest answering but indirectly provide totally irrelevant information. 

\section*{Ethical considerations}
This work studies LLM \emph{robustness} to authority and template-style framing, rather than the construction of new jailbreak techniques. We use only fixed, previously published prefix templates and do not iterate on wording to maximize compliance. Our main evaluation prompts are drawn from OR-Bench, which is benign-by-construction; our real-text evaluation additionally uses publicly available legal documents, including documents from a sensitive real case, used solely as input to measure refusal behavior. Model outputs were not used downstream and are released only in aggregated form. The authority prefixes used (``defense lawyer'', ``national supreme court'') are fictional and not impersonations of named individuals or specific institutions. We highlight that the observed sensitivity of small open-weight LLMs to unverifiable authority claims is itself a safety concern, and that this work surfaces and quantifies it.

\section*{AI Usage Statement}
We used Anthropic's Claude (Opus~4.6, Opus~4.7, and Opus~5) as a coding and
research assistant, for code scaffolding, debugging the experimental pipeline
(Ollama client, CSV manipulation), and for drafting portions of this manuscript.
All experimental design choices, the selection of evaluated models, statistical
interpretation, and final manuscript wording were made by the authors. No model
outputs were used as data in any results table. As such, authors claim full responsibility of the presented findings and the final text submitted.

\bibliography{custom}

@misc{GoogleGeminiPromptGuide2024,
  title = {Gemini for {Google} {Workspace}: Prompting Guide 101 --- {A} quick-start handbook for effective prompts},
  author = {{Google}},
  year = {2024},
  month = {October},
  url = {https://services.google.com/fh/files/misc/gemini_for_workspace_prompt_guide_october_2024_digital_final.pdf},
  howpublished = {Online PDF}
}

@misc{DeepL,
  title = {{DeepL} {Translator}},
  author = {{DeepL SE}},
  year = {2024},
  url = {https://www.deepl.com/translator},
  howpublished = {Online translation service}
}

@misc{CLOUD2018,
  title={Clarifying Lawful Overseas Use of Data Act},
  author={{United States Congress}},
  year={2018},
  note={Pub. L. 115-141, 18 U.S.C. § 2713},
  url={https://www.congress.gov/115/plaws/publ141/PLAW-115publ141.htm}
}

@inproceedings{cui2025orbench,
  title = {{OR-Bench}: An Over-Refusal Benchmark for Large Language Models},
  author = {Cui, Justin and Chiang, Wei-Lin and Stoica, Ion and Hsieh, Cho-Jui},
  booktitle = {Proceedings of the 42nd International Conference on Machine Learning (ICML)},
  series = {PMLR},
  volume = {267},
  pages = {11515--11542},
  year = {2025}
}

@inproceedings{rottger2024xstest,
  title = "{XST}est: A Test Suite for Identifying Exaggerated Safety Behaviours in Large Language Models",
    author = {R{\"o}ttger, Paul  and
      Kirk, Hannah  and
      Vidgen, Bertie  and
      Attanasio, Giuseppe  and
      Bianchi, Federico  and
      Hovy, Dirk},
    editor = "Duh, Kevin  and
      Gomez, Helena  and
      Bethard, Steven",
    booktitle = "Proceedings of the 2024 Conference of the North American Chapter of the Association for Computational Linguistics: Human Language Technologies (Volume 1: Long Papers)",
    month = jun,
    year = "2024",
    address = "Mexico City, Mexico",
    publisher = "Association for Computational Linguistics",
    url = "https://aclanthology.org/2024.naacl-long.301/",
    doi = "10.18653/v1/2024.naacl-long.301",
    pages = "5377--5400"
}

@inproceedings{cheng2025overt,
  title = {{OVERT}: A Benchmark for Over-Refusal Evaluation on Text-to-Image Models},
  author = {Cheng, Ziheng and Huang, Yixiao and Xu, Hui and Sojoudi, Somayeh and Zhao, Xuandong and Song, Dawn and Mei, Song},
  booktitle = {Advances in Neural Information Processing Systems (NeurIPS)},
  year = {2025}
}

@article{xue2026refusal,
  title={Deactivating Refusal Triggers: Understanding and Mitigating Overrefusal in Safety Alignment}, 
      author={Zhiyu Xue and Zimo Qi and Guangliang Liu and Bocheng Chen and Ramtin Pedarsani},
      year={2026},
      eprint={2603.11388},
      archivePrefix={arXiv},
      primaryClass={cs.AI},
      url={https://arxiv.org/abs/2603.11388}, 
}

@article{campbell2026defensive,
  title={Defensive Refusal Bias: How Safety Alignment Fails Cyber Defenders}, 
      author={David Campbell and Neil Kale and Udari Madhushani Sehwag and Bert Herring and Nick Price and Dan Borges and Alex Levinson and Christina Q Knight},
      year={2026},
      eprint={2603.01246},
      archivePrefix={arXiv},
      primaryClass={cs.CR},
      url={https://arxiv.org/abs/2603.01246},
}

@article{dabas2025actor,
  title={Just Enough Shifts: Mitigating Over-Refusal in Aligned Language Models with Targeted Representation Fine-Tuning}, 
      author={Mahavir Dabas and Si Chen and Charles Fleming and Ming Jin and Ruoxi Jia},
      year={2025},
      eprint={2507.04250},
      archivePrefix={arXiv},
      primaryClass={cs.LG},
      url={https://arxiv.org/abs/2507.04250}, 
}

@article{fei2024lawbench,
  title = {{LawBench}: Benchmarking Legal Knowledge of Large Language Models},
  author = {Fei, Zhiwei and Shen, Xiaoyu and Zhu, Dawei and Zhou, Fengzhe and Han, Zhuo and Zhang, Songyang and Chen, Kai and Shen, Zongwen and Ge, Jidong},
  journal = {arXiv preprint arXiv:2309.16289},
  year = {2023}
}

@inproceedings{li2024lexeval,
  title={LexEval: A Comprehensive Chinese Legal Benchmark for Evaluating Large Language Models}, 
      author={Haitao Li and You Chen and Qingyao Ai and Yueyue Wu and Ruizhe Zhang and Yiqun Liu},
      year={2024},
      eprint={2409.20288},
      archivePrefix={arXiv},
      primaryClass={cs.CL},
      url={https://arxiv.org/abs/2409.20288}, 
}

@article{cao2025safelawbench,
  title = "{S}afe{L}aw{B}ench: Towards Safe Alignment of Large Language Models",
    author = "Cao, Chuxue  and
      Zhu, Han  and
      Ji, Jiaming  and
      Sun, Qichao  and
      Zhu, Zhenghao  and
      Yinyu, Wu  and
      Dai, Josef  and
      Yang, Yaodong  and
      Han, Sirui  and
      Guo, Yike",
    editor = "Che, Wanxiang  and
      Nabende, Joyce  and
      Shutova, Ekaterina  and
      Pilehvar, Mohammad Taher",
    booktitle = "Findings of the Association for Computational Linguistics: ACL 2025",
    month = jul,
    year = "2025",
    address = "Vienna, Austria",
    publisher = "Association for Computational Linguistics",
    url = "https://aclanthology.org/2025.findings-acl.721/",
    doi = "10.18653/v1/2025.findings-acl.721",
    pages = "14015--14048",
    ISBN = "979-8-89176-256-5"
}

@inproceedings{LegalBERT2020,
  author       = {Ilias Chalkidis and
                  Manos Fergadiotis and
                  Prodromos Malakasiotis and
                  Nikolaos Aletras and
                  Ion Androutsopoulos},
  editor       = {Trevor Cohn and
                  Yulan He and
                  Yang Liu},
  title        = {{LEGAL-BERT:} The Muppets straight out of Law School},
  booktitle    = {Findings of the Association for Computational Linguistics: {EMNLP}
                  2020, Online Event, 16-20 November 2020},
  series       = {Findings of {ACL}},
  pages        = {2898--2904},
  publisher    = {Association for Computational Linguistics},
  year         = {2020},
  url          = {https://doi.org/10.18653/v1/2020.findings-emnlp.261},
  doi          = {10.18653/V1/2020.FINDINGS-EMNLP.261},
  bibsource    = {dblp computer science bibliography, https://dblp.org}
}

@article{LawFormer2021,
  author       = {Chaojun Xiao and
                  Xueyu Hu and
                  Zhiyuan Liu and
                  Cunchao Tu and
                  Maosong Sun},
  title        = {Lawformer: {A} pre-trained language model for Chinese legal long documents},
  journal      = {{AI} Open},
  volume       = {2},
  pages        = {79--84},
  year         = {2021},
  url          = {https://doi.org/10.1016/j.aiopen.2021.06.003},
  doi          = {10.1016/J.AIOPEN.2021.06.003},
  bibsource    = {dblp computer science bibliography, https://dblp.org}
}

@article{AIandLawReview2012,
  author       = {Trevor J. M. Bench{-}Capon and
                  Michal Araszkiewicz and
                  Kevin D. Ashley and
                  Katie Atkinson and
                  Floris Bex and
                  Filipe Borges and
                  Dani{\`{e}}le Bourcier and
                  Paul Bourgine and
                  Jack G. Conrad and
                  Enrico Francesconi and
                  Thomas F. Gordon and
                  Guido Governatori and
                  Jochen L. Leidner and
                  David D. Lewis and
                  Ronald Prescott Loui and
                  L. Thorne McCarty and
                  Henry Prakken and
                  Frank Schilder and
                  Erich Schweighofer and
                  Paul Thompson and
                  Alex Tyrrell and
                  Bart Verheij and
                  Douglas N. Walton and
                  Adam Z. Wyner},
  title        = {A history of {AI} and Law in 50 papers: 25 years of the international
                  conference on {AI} and Law},
  journal      = {Artif. Intell. Law},
  volume       = {20},
  number       = {3},
  pages        = {215--319},
  year         = {2012},
  url          = {https://doi.org/10.1007/s10506-012-9131-x},
  doi          = {10.1007/S10506-012-9131-X},
  bibsource    = {dblp computer science bibliography, https://dblp.org}
}

@inproceedings{LegalBench2023,
  author       = {Neel Guha and
                  Julian Nyarko and
                  Daniel E. Ho and
                  Christopher R{\'{e}} and
                  Adam Chilton and
                  K. Aditya and
                  Alex Chohlas{-}Wood and
                  Austin Peters and
                  Brandon Waldon and
                  Daniel N. Rockmore and
                  Diego Zambrano and
                  Dmitry Talisman and
                  Enam Hoque and
                  Faiz Surani and
                  Frank Fagan and
                  Galit Sarfaty and
                  Gregory M. Dickinson and
                  Haggai Porat and
                  Jason Hegland and
                  Jessica Wu and
                  Joe Nudell and
                  Joel Niklaus and
                  John J. Nay and
                  Jonathan H. Choi and
                  Kevin Tobia and
                  Margaret Hagan and
                  Megan Ma and
                  Michael A. Livermore and
                  Nikon Rasumov{-}Rahe and
                  Nils Holzenberger and
                  Noam Kolt and
                  Peter Henderson and
                  Sean Rehaag and
                  Sharad Goel and
                  Shang Gao and
                  Spencer Williams and
                  Sunny Gandhi and
                  Tom Zur and
                  Varun Iyer and
                  Zehua Li},
  editor       = {Alice Oh and
                  Tristan Naumann and
                  Amir Globerson and
                  Kate Saenko and
                  Moritz Hardt and
                  Sergey Levine},
  title        = {LegalBench: {A} Collaboratively Built Benchmark for Measuring Legal
                  Reasoning in Large Language Models},
  booktitle    = {Advances in Neural Information Processing Systems 36: Annual Conference
                  on Neural Information Processing Systems 2023, NeurIPS 2023, New Orleans,
                  LA, USA, December 10 - 16, 2023},
  year         = {2023},
  url          = {http://papers.nips.cc/paper\_files/paper/2023/hash/89e44582fd28ddfea1ea4dcb0ebbf4b0-Abstract-Datasets\_and\_Benchmarks.html},
  bibsource    = {dblp computer science bibliography, https://dblp.org}
}

@article{GPT4PassesBarExam2024,
    author = {Katz, Daniel Martin and Bommarito, Michael James and Gao, Shang and Arredondo, Pablo},
    title = {GPT-4 passes the bar exam},
    journal = {Philosophical Transactions of the Royal Society A: Mathematical, Physical and Engineering Sciences},
    volume = {382},
    number = {2270},
    pages = {20230254},
    year = {2024},
    month = {02},
    issn = {1364-503X},
    doi = {10.1098/rsta.2023.0254},
    url = {https://doi.org/10.1098/rsta.2023.0254},
    eprint = {https://royalsocietypublishing.org/rsta/article-pdf/doi/10.1098/rsta.2023.0254/1328474/rsta.2023.0254.pdf},
}

@article{GPT4DoesNotPassBar2025,
  author       = {Eric Mart{\'{\i}}nez},
  title        = {Re-evaluating GPT-4's bar exam performance},
  journal      = {Artif. Intell. Law},
  volume       = {33},
  number       = {3},
  pages        = {581--604},
  year         = {2025},
  url          = {https://doi.org/10.1007/s10506-024-09396-9},
  doi          = {10.1007/S10506-024-09396-9},
  bibsource    = {dblp computer science bibliography, https://dblp.org}
}

@article{LegalHallucination2024,
  author       = {Varun Magesh and
                  Faiz Surani and
                  Matthew Dahl and
                  Mirac Suzgun and
                  Christopher D. Manning and
                  Daniel E. Ho},
  title        = {Hallucination-Free? Assessing the Reliability of Leading {AI} Legal
                  Research Tools},
  journal      = {CoRR},
  volume       = {abs/2405.20362},
  year         = {2024},
  url          = {https://doi.org/10.48550/arXiv.2405.20362},
  doi          = {10.48550/ARXIV.2405.20362},
  eprinttype   = {arXiv},
  eprint       = {2405.20362},
  bibsource    = {dblp computer science bibliography, https://dblp.org}
}

@article{LargeLegalFiction2024,
  author       = {Matthew Dahl and
                  Varun Magesh and
                  Mirac Suzgun and
                  Daniel E. Ho},
  title        = {Large Legal Fictions: Profiling Legal Hallucinations in Large Language
                  Models},
  journal      = {CoRR},
  volume       = {abs/2401.01301},
  year         = {2024},
  url          = {https://doi.org/10.48550/arXiv.2401.01301},
  doi          = {10.48550/ARXIV.2401.01301},
  eprinttype   = {arXiv},
  eprint       = {2401.01301},
  bibsource    = {dblp computer science bibliography, https://dblp.org}
}

@article{GPT4Report2023,
  author       = {OpenAI},
  title        = {{GPT-4} Technical Report},
  journal      = {CoRR},
  volume       = {abs/2303.08774},
  year         = {2023},
  url          = {https://doi.org/10.48550/arXiv.2303.08774},
  doi          = {10.48550/ARXIV.2303.08774},
  eprinttype   = {arXiv},
  eprint       = {2303.08774},
  bibsource    = {dblp computer science bibliography, https://dblp.org}
}

@article{Dehghani2025LargeLM,
  title={Large Language Models in Legal Systems: A Survey},
  author={Fatemeh Dehghani and Roya Dehghani and Yazdan Naderzadeh Ardebili and Shahryar Rahnamayan},
  journal={Humanities and Social Sciences Communications},
  year={2025},
  volume={12},
  url={https://api.semanticscholar.org/CorpusID:284295301}
}

@article{LLMsLawSurvey2024,
  author       = {Jinqi Lai and
                  Wensheng Gan and
                  Jiayang Wu and
                  Zhenlian Qi and
                  Philip S. Yu},
  title        = {Large language models in law: {A} survey},
  journal      = {{AI} Open},
  volume       = {5},
  pages        = {181--196},
  year         = {2024},
  url          = {https://doi.org/10.1016/j.aiopen.2024.09.002},
  doi          = {10.1016/J.AIOPEN.2024.09.002},
  bibsource    = {dblp computer science bibliography, https://dblp.org}
}

@incollection{usingOllama2025,
  author = {Marcondes, Francisco and Gala, Adelino and Magalhães, Renata and Britto, Fernando and Duraes, Dalila and Novais, Paulo},
year = {2025},
month = {02},
pages = {23-35},
title = {Using Ollama},
isbn = {978-3-031-76630-5},
doi = {10.1007/978-3-031-76631-2_3}
}

@software{mlx2023,
  author = {Awni Hannun and Jagrit Digani and Angelos Katharopoulos and Ronan Collobert},
  title = {{MLX}: Efficient and flexible machine learning on Apple silicon},
  url = {https://github.com/ml-explore},
  version = {0.0},
  year = {2023},
}

@article{Apertus2025,
  author       = {Project Apertus},
  title        = {Apertus: Democratizing Open and Compliant LLMs for Global Language
                  Environments},
  journal      = {CoRR},
  volume       = {abs/2509.14233},
  year         = {2025},
  url          = {https://doi.org/10.48550/arXiv.2509.14233},
  doi          = {10.48550/ARXIV.2509.14233},
  eprinttype   = {arXiv},
  eprint       = {2509.14233},
  bibsource    = {dblp computer science bibliography, https://dblp.org}
}

@book{UNESCO2025Guidelines,
  author = {Gutiérrez, Juan David and {UNESCO}},
  title = {Guidelines for the use of {AI} systems in courts and tribunals},
  publisher = {UNESCO},
  year = {2025},
  isbn = {978-92-3-100830-6},
  doi = {10.58338/LIEY8089},
  pages = {44},
  url = {https://doi.org/10.58338/LIEY8089}
}

@misc{yang2025qwen3technicalreport,
      title={Qwen3 Technical Report}, 
      author={An Yang and Anfeng Li and Baosong Yang and Beichen Zhang and Binyuan Hui and Bo Zheng and Bowen Yu and Chang Gao and Chengen Huang and Chenxu Lv and Chujie Zheng and Dayiheng Liu and Fan Zhou and Fei Huang and Feng Hu and Hao Ge and Haoran Wei and Huan Lin and Jialong Tang and Jian Yang and Jianhong Tu and Jianwei Zhang and Jianxin Yang and Jiaxi Yang and Jing Zhou and Jingren Zhou and Junyang Lin and Kai Dang and Keqin Bao and Kexin Yang and Le Yu and Lianghao Deng and Mei Li and Mingfeng Xue and Mingze Li and Pei Zhang and Peng Wang and Qin Zhu and Rui Men and Ruize Gao and Shixuan Liu and Shuang Luo and Tianhao Li and Tianyi Tang and Wenbiao Yin and Xingzhang Ren and Xinyu Wang and Xinyu Zhang and Xuancheng Ren and Yang Fan and Yang Su and Yichang Zhang and Yinger Zhang and Yu Wan and Yuqiong Liu and Zekun Wang and Zeyu Cui and Zhenru Zhang and Zhipeng Zhou and Zihan Qiu},
      year={2025},
      eprint={2505.09388},
      archivePrefix={arXiv},
      primaryClass={cs.CL},
      url={https://arxiv.org/abs/2505.09388}, 
}

@article{LLaMA3Herd2024,
  author       = {Llama Team},
  title        = {The Llama 3 Herd of Models},
  journal      = {CoRR},
  volume       = {abs/2407.21783},
  year         = {2024},
  url          = {https://doi.org/10.48550/arXiv.2407.21783},
  doi          = {10.48550/ARXIV.2407.21783},
  eprinttype   = {arXiv},
  eprint       = {2407.21783},
  bibsource    = {dblp computer science bibliography, https://dblp.org}
}

@article{Gemma4,
  author       = {{Gemma Team}},
  title        = {Gemma 4 Technical Report},
  journal      = {CoRR},
  volume       = {abs/2607.02770},
  year         = {2026},
  url          = {https://doi.org/10.48550/arXiv.2607.02770},
  doi          = {10.48550/ARXIV.2607.02770},
  eprinttype   = {arXiv},
  eprint       = {2607.02770},
  primaryClass = {cs.CL}
}

@article{EmpiricalJailbreaks2023,
  author       = {Yi Liu and
                  Gelei Deng and
                  Zhengzi Xu and
                  Yuekang Li and
                  Yaowen Zheng and
                  Ying Zhang and
                  Lida Zhao and
                  Tianwei Zhang and
                  Yang Liu},
  title        = {Jailbreaking ChatGPT via Prompt Engineering: An Empirical Study},
  journal      = {CoRR},
  volume       = {abs/2305.13860},
  year         = {2023},
  url          = {https://doi.org/10.48550/arXiv.2305.13860},
  doi          = {10.48550/ARXIV.2305.13860},
  eprinttype   = {arXiv},
  eprint       = {2305.13860},
  bibsource    = {dblp computer science bibliography, https://dblp.org}
}

@inproceedings{chao2024jailbreakbench,
  title = {{JailbreakBench}: An Open Robustness Benchmark for Jailbreaking Large Language Models},
  author = {Chao, Patrick and Debenedetti, Edoardo and Robey, Alexander and Andriushchenko, Maksym and Croce, Francesco and Sehwag, Vikash and Dobriban, Edgar and Flammarion, Nicolas and Pappas, George J. and Tram{\`e}r, Florian and Hassani, Hamed and Wong, Eric},
  booktitle = {Advances in Neural Information Processing Systems (NeurIPS), Datasets and Benchmarks Track},
  year = {2024}
}

@misc{bger2026,
  author       = {{Swiss Federal Supreme Court}},
  title        = {{Tribunal f\'ed\'eral / Schweizerisches Bundesgericht / Tribunale federale}},
  howpublished = {\url{https://www.bger.ch/fr/index.htm}},
  year         = {2026},
  note         = {Accessed: 2026-05-25}
}

@article{zou2023gcg,
  title         = {Universal and Transferable Adversarial Attacks on Aligned Language Models},
  author        = {Zou, Andy and Wang, Zifan and Carlini, Nicholas and Nasr, Milad and
                   Kolter, J. Zico and Fredrikson, Matt},
  year          = {2023},
  eprint        = {2307.15043},
  archivePrefix = {arXiv},
  primaryClass  = {cs.CL},
  url           = {https://arxiv.org/abs/2307.15043}
}

@misc{widener2024legalpatterns,
  title = {Legal Prompt Patterns},
  author = {{Widener University Delaware Law School Library}},
  howpublished = {LibGuides: Generative AI and Legal Research},
  year = {2024},
  note = {\url{https://libguides.law.widener.edu/c.php?g=1342893&p=10038411}}
}

@techreport{efk2026bger,
  author      = {{Swiss Federal Audit Office (EFK)}},
  title        = {Audit of the Use of Artificial Intelligence: Federal Supreme Court},
  institution = {Swiss Federal Audit Office (Eidgen\"ossische Finanzkontrolle)},
  number      = {EFK-25732},
  year        = {2026},
  month       = jan,
  note        = {Version incl. statements, 23 January 2026},
  url         = {https://www.efk.admin.ch/wp-content/uploads/publikationen/berichte/wirtschaft_und_verwaltung/informatikprojekte/25732/25732be-bger-einsatz-kunstlicher-intelligenz.pdf}
}

@article{Qwen25TechnicalRep2025,
  author       = {An Yang and
                  Baosong Yang and
                  Beichen Zhang and
                  Binyuan Hui and
                  Bo Zheng and
                  Bowen Yu and
                  Chengyuan Li and
                  Dayiheng Liu and
                  Fei Huang and
                  Haoran Wei and
                  Huan Lin and
                  Jian Yang and
                  Jianhong Tu and
                  Jianwei Zhang and
                  Jianxin Yang and
                  Jiaxi Yang and
                  Jingren Zhou and
                  Junyang Lin and
                  Kai Dang and
                  Keming Lu and
                  Keqin Bao and
                  Kexin Yang and
                  Le Yu and
                  Mei Li and
                  Mingfeng Xue and
                  Pei Zhang and
                  Qin Zhu and
                  Rui Men and
                  Runji Lin and
                  Tianhao Li and
                  Tingyu Xia and
                  Xingzhang Ren and
                  Xuancheng Ren and
                  Yang Fan and
                  Yang Su and
                  Yichang Zhang and
                  Yu Wan and
                  Yuqiong Liu and
                  Zeyu Cui and
                  Zhenru Zhang and
                  Zihan Qiu},
  title        = {Qwen2.5 Technical Report},
  journal      = {CoRR},
  volume       = {abs/2412.15115},
  year         = {2024},
  url          = {https://doi.org/10.48550/arXiv.2412.15115},
  doi          = {10.48550/ARXIV.2412.15115},
  eprinttype   = {arXiv},
  eprint       = {2412.15115},
  bibsource    = {dblp computer science bibliography, https://dblp.org}
}

\appendix

\section{Robustness to prefix wording.}\label{sec:appendix_prefix_formulation}
Ideally, to rule out confounders such as prefix length or formality, one would test many different phrasings per condition, keeping the underlying request fixed and varying only the legal authority explanation. As a first verification that the effect is not an artifact of reformulation, we constructed paraphrases of the lawyer and supreme-court prefixes, matched in length and style and differing only in wording, and re-tested them on the most refusal-prone English category (\emph{sexual}). Where tested, the paraphrases reproduce the effect and keep
refusals far above the no-prefix baseline. For Llama the paraphrased and original prefixes give almost identical counts (lawyer 11 vs.\ 12, supreme-court 15 vs.\ 16 out of 200), and for Apertus the paraphrases yield even higher refusal (lawyer 34 vs.\ 57, supreme-court 29 vs.\ 55). Absolute magnitudes vary
somewhat with wording (e.g.\ Gemma supreme-court 43 vs.\ 34), but the direction of the authority effect is unchanged, indicating it is driven by the authority claim itself rather than the surface form of the prompt. Details are available in
the repository.\footnote{\url{https://anonymous.4open.science/r/Overrefusal_in_Criminal_Legal_Context-DB01/}}
%
%
%

\section{Statistical Analysis}
\label{app:stats}

For all the experiments, we pair every prompt's baseline response (no framing
prefix) with its response under a legal-authority/jailbreak framing
prefix, and run McNemar's exact test on the
pairs. It is one-sided: does the prefix \emph{increase} refusals?  For a given model and prefix, it considers only the prompts whose refusal label \emph{changes} between the no-prefix baseline and the prefixed condition and reports the odds ratio that evaluates whether the prefix pushes more prompts into refusal than out of it. 

Then, to avoid false-positive conclusions just because of the multiple tests, we apply Holm-Bonferroni correction. For the OR-Bench dataset correction runs on 20 tests, for BGER - on 12.  This procedure sorts the $p$-values and applies a progressively less strict threshold, so that an effect declared significant is unlikely to be an artifact. Our effects are large enough and remain significant after the correction.

\subsection{Aggregation tables }

Full per-cell statistics are reported in Table~\ref{tab:mcnemar_orbench} (OR-Bench) and Table~\ref{tab:mcnemar_bger} (real BGer paragraphs). We see the conclusion in the column ``Significant'' ---
all the experiments for Llama, Apertus and Gemma on all three languages and
both authority prefixes show significant results on the hypothesis ``prefix
increases refusal rate''.

\begin{table}[!ht]
\centering
\footnotesize
\setlength{\tabcolsep}{4pt}
\renewcommand{\arraystretch}{0.95}
\begin{tabular}{lllrrc}
\toprule
Model & Lang & Prefix & None$\to$Pref & $p$ (Holm) & Sig. \\
\midrule
Llama   & en & law.  & 7.2$\to$11.1  & $8.8\times10^{-4}$   & Yes \\
Llama   & en & sup.  & 7.3$\to$13.2  & $8.1\times10^{-7}$   & Yes \\
Llama   & en & jail. & 7.3$\to$27.5  & $6.3\times10^{-51}$  & Yes \\
Llama   & fr & sup.  & 16.8$\to$46.9 & $1.6\times10^{-85}$  & Yes \\
Llama   & de & sup.  & 16.5$\to$53.3 & $1.6\times10^{-100}$ & Yes \\
Gemma   & en & law.  & 10.7$\to$15.2 & $6.3\times10^{-4}$   & Yes \\
Gemma   & en & sup.  & 10.7$\to$21.8 & $1.1\times10^{-15}$  & Yes \\
Gemma   & en & jail. & 10.7$\to$9.3  & $9.4\times10^{-1}$   & No  \\
Gemma   & fr & sup.  & 6.3$\to$12.7  & $7.7\times10^{-7}$   & Yes \\
Gemma   & de & sup.  & 4.0$\to$7.4   & $3.7\times10^{-4}$   & Yes \\
Apertus & en & law.  & 5.6$\to$14.3  & $7.3\times10^{-18}$  & Yes \\
Apertus & en & sup.  & 5.6$\to$12.2  & $2.3\times10^{-12}$  & Yes \\
Apertus & en & jail. & 5.6$\to$9.5   & $9.1\times10^{-6}$   & Yes \\
Apertus & fr & sup.  & 1.4$\to$25.1  & $3.1\times10^{-65}$  & Yes \\
Apertus & de & sup.  & 1.9$\to$3.5   & $1.1\times10^{-2}$   & Yes \\
Qwen    & en & law.  & 2.2$\to$3.8   & $4.5\times10^{-2}$   & Yes \\
Qwen    & en & sup.  & 2.2$\to$2.6   & $6.3\times10^{-1}$   & No  \\
Qwen    & en & jail. & 2.2$\to$3.1   & $2.6\times10^{-1}$   & No  \\
Qwen    & fr & sup.  & 0.0$\to$0.2   & $2.5\times10^{-1}$   & No  \\
Qwen    & de & sup.  & 0.0$\to$0.3   & $2.5\times10^{-1}$   & No  \\
\bottomrule
\end{tabular}
\caption{OR-Bench: paired refusal rates (\%) under authority/jailbreak framing
vs.\ the \texttt{none} baseline. Models are the 8B variants (Gemma: 4~E4B).
Prefixes: \emph{law.}\ lawyer, \emph{sup.}\ supreme-court, \emph{jail.}\
jailbreak. None$\to$Pref gives baseline and prefixed refusal rates.
$p$~(Holm) is the McNemar exact-test $p$-value after Holm-Bonferroni correction
for multiple comparisons; \emph{Sig.}\ marks whether the increase is significant
at $\alpha=0.05$.}
\label{tab:mcnemar_orbench}
\end{table}

\begin{table}[!ht]
\centering
\footnotesize
\setlength{\tabcolsep}{4pt}
\renewcommand{\arraystretch}{0.95}
\begin{tabular}{llrrc}
\toprule
Model & Lang & None$\to$Pref & $p$ (Holm) & Sig. \\
\midrule
Llama   & en & 18.8$\to$68.9 & $3.9\times10^{-249}$ & Yes \\
Llama   & fr & 12.5$\to$23.4 & $2.3\times10^{-40}$  & Yes \\
Llama   & de & 25.6$\to$28.5 & $1.8\times10^{-2}$   & Yes \\
Apertus & fr & 1.1$\to$19.3  & $4.3\times10^{-76}$  & Yes \\
Apertus & en & 1.4$\to$3.9   & $7.0\times10^{-6}$   & Yes \\
Apertus & de & 1.0$\to$2.5   & $1.1\times10^{-3}$   & Yes \\
Gemma   & en & 0.5$\to$0.1   & $1.0\times10^{0}$    & No  \\
Gemma   & fr & 0.4$\to$0.1   & $1.0\times10^{0}$    & No  \\
Gemma   & de & 0.2$\to$0.1   & $1.0\times10^{0}$    & No  \\
Qwen    & de & 0.1$\to$0.4   & $2.2\times10^{-1}$   & No  \\
Qwen    & en & 0.2$\to$0.2   & $1.0\times10^{0}$    & No  \\
Qwen    & fr & 0.0$\to$0.1   & $1.0\times10^{0}$    & No  \\
\bottomrule
\end{tabular}
\caption{Real BGer (Swiss federal court) paragraphs: paired refusal rates (\%)
under the supreme-court prefix vs.\ the \texttt{none} baseline. None$\to$Pref
gives baseline and prefixed rates. $p$~(Holm) is the McNemar exact-test
$p$-value after Holm-Bonferroni correction; \emph{Sig.}\ marks significance at
$\alpha=0.05$.}
\label{tab:mcnemar_bger}
\end{table}

Full per-cell statistics for every experiment in the paper (the counts $b$ and
$c$, the odds ratio, baseline and prefixed refusal rates, and both uncorrected
and corrected $p$-values) are released with our code.\footnote{\url{https://anonymous.4open.science/r/Overrefusal_in_Criminal_Legal_Context-DB01/}}

\end{document}